# Fault Analysis And Predictive Maintenance Of Induction Motor Using Machine Learning.

Kavana V
*Department of Electrical and Electronics engineering*
*JSS Science and Technology University*
Mysuru, India.
kavanavvasishta1863@gmail.com

Neethi M
*Department of Electrical and Electronics engineering*
*JSS Science and Technology University*
Mysuru, India.
neethisjce@gmail.com

*Abstract*

**Induction motors are one of the most crucial electrical equipment and are extensively used in industries in a wide range of applications. This paper presents a machine learning model for the fault detection and classification of induction motor faults by using three phase voltages and currents as inputs. The aim of this work is to protect vital electrical components and to prevent abnormal event progression through early detection and diagnosis. This work presents a fast forward artificial neural network model to detect some of the commonly occurring electrical faults like overvoltage, under voltage, single phasing, unbalanced voltage, overload, ground fault. A separate model free monitoring system wherein the motor itself acts like a sensor is presented and the only monitored signals are the input given to the motor. Limits for current and voltage values are set for the faulty and healthy conditions, which is done by a classifier. Real time data from a 0.33 HP induction motor is used to train and test the neural network. The model so developed analyses the voltage and current values given at a particular instant and classifies the data into no fault or the specific fault. The model is then interfaced with a real motor to accurately detect and classify the faults so that further necessary action can be taken.**

I. INTRODUCTION

Induction motors are the most extensively used electrical motors because of their simple construction, ruggedness and low cost. More than 90% of the industries use induction motors, mostly as electrical drives since they can be designed for a wide range of power ratings. In spite of their flexibility and robustness, they are subjected to many catastrophic failures. Identifying these faults at an early stage and attending to them is very important as otherwise they lead to huge production and financial losses. Pre- fault detection and isolation of the healthy parts also prevents fault progression and failure of other more vital components. Industries use a large number of motors and hence, their manual maintenance is tedious and unreliable. Therefore, many attempts have been made towards automatic maintenance. Earlier, conditional monitoring of electrical machines was employed and the electromechanical relays were put to use to achieve this. But these relays are slow in operation and also lead to huge power losses due to mechanical parts involved [6]. Hence, they cannot be used in critical applications requiring small response times. Electromechanical relays were later replaced by solid state relays as they consume very less power and are comparatively fast. With the advent of microprocessors, attempts were made for conditional monitoring of machines by using pre- written programmes downloaded onto the microprocessor chips.

The above mentioned techniques, however could not guarantee maximum safety and reliability as they cannot address catastrophic failures. Computer revolution with the dawn of machine learning drew the attention of scientists and they began to think of ways through which these techniques could be used to monitor and safeguard machines. Machine learning models take the place of human to intelligently monitor and maintain the specified system tasks [4]. Artificial neural networks are very handy in this regard as they can handle huge amount of data, have small response time and can effectively handle non- linearity (which most of the time is the inherent characteristic of electromechanical systems) [10].

The aim of this work is to prevent fault progression and protect vital components of the power system by early detection of electrical faults of three phase induction motors using artificial neural network. We have addressed seven classes of electrical faults of induction motors; overload, ground fault, locked rotor, single phasing, over voltage, under voltage and unbalanced supply voltage.

Section II provides brief information about the adopted neural network methodology. Results of the work are discussed in section III. Real system integration is explained in section IV. Section V gives a comprehensive view of similar machine learning based techniques while section VI offers future scope. The findings are again analysed in section VII.

## II. ARTIFICIAL NEURAL NETWORK APPROACH.

The ANN comes under machine learning approaches. Any machine learning problem is solved through the following steps.

1. Defining objectives: The first step towards addressing any machine learning problem is to define the goals. The problem might be to find a numerical output value (linear regression), to separate into classes (classification) or clustering. The type of machine learning approach that needs to be used is decided by studying the problem and extracting the objectives out of it.
2. Data acquisition: The data required to train the neural network is obtained in real time. The type and amount of data required depends on the application.
3. Separating train and test data: The collected data is then divided into train and test data to train the neural network model and then test its working and calculate its accuracy [12].
4. Analysing the findings: The obtained results are then analysed to check for any errors and also to identify scope for improvement, if any.
5. Interfacing with real system: Once the proper working of the trained model is confirmed, it is interfaced with a real system and is then used in real time applications.

The Artificial Neural Network (ANN) is designed to mimic human brain to think for itself and take action without being explicitly programmed. Just like a human brain, the ANN model consists of artificial neurons or nodes as the basic building blocks. The nodes are arranged in layers: input layer, output layer and one or more hidden layers. The hidden layer is used to increase the accuracy of classification. The working of the algorithm can be explained from the below steps.

1) The algorithm assigns weights or priorities to each of the input variable.
2) It then multiplies the weight with the input variable, $x_i$ and adds bias to it. The same thing is done to each of the node and the summation of these values is transmitted to the next layer.
The learning proceeds as data moves through each layer.
3) The data at the end of output node is then compared with a set threshold value and if the calculated value exceeds the threshold, the input vector is classified to the intended class or else is passed over to check for other classes.

The block diagram of the working of a feed forward back loop neural network algorithm is as shown in fig.1.

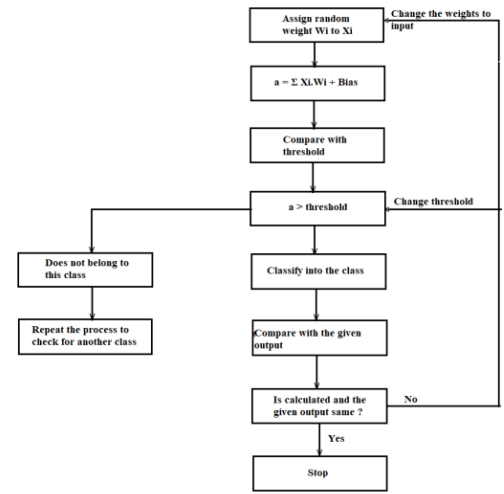

Fig 1: Block diagram of ANN algorithm

4) The calculated class is compared with the actual output and any error is corrected by varying the number of layers and the number of neurons.
5) The process is continued till an accurate fit is obtained and the model is hence, trained. The equation of the algorithm is,

$$\Sigma W_i X_i + Bias \geq Threshold$$

In this work, three phase voltages and currents of a three phase 1/3 HP, 208 V induction motor is used as the input. Data is collected in real time for each of the classes and is used to train and test the neural network. MATLAB is the software used whose neural network toolboxes train and tests the data.

## III. RESULTS AND DISCUSSION

In this work, the artificial neural network model is created and trained by using three phase voltages and currents as the input. The model is trained for each of the seven electrical faults using approximately 800 value sets of three phase voltages and currents. A sample of the used database is as shown in table 1.

Table 1: Sample database

| CLASS | $V_1$ | $V_2$ | $V_3$ | $I_1$ | $I_2$ | $I_3$ |
|---|---|---|---|---|---|---|
| 1 | 2.661025 | 2.624276 | 2.701274 | 0.490768 | 0.478549 | 0.493368 |
| 1 | 2.660319 | 2.624661 | 2.700700 | 0.491114 | 0.478722 | 0.492584 |
| 2 | 2.647625 | 2.598815 | 2.671626 | 0.006194 | 0.643518 | 0.640217 |
| 2 | 2.650816 | 2.601661 | 2.673722 | 0.006123 | 0.641548 | 0.638553 |
| 3 | 0.919570 | 2.621412 | 2.626511 | 0.172113 | 0.772419 | 0.662758 |
| 3 | 0.919852 | 2.621627 | 2.625342 | 0.172072 | 0.772106 | 0.662653 |
| 4 | 1.874796 | 1.855089 | 1.874878 | 0.286777 | 0.287052 | 0.281013 |
| 4 | 1.452803 | 1.449902 | 1.441935 | 0.245231 | 0.249739 | 0.234152 |
| 5 | 2.865128 | 2.871906 | 2.855436 | 0.482896 | 0.499206 | 0.496894 |
| 5 | 2.868791 | 2.875877 | 2.860353 | 0.483453 | 0.499363 | 0.496879 |
| 6 | 2.657179 | 2.613409 | 2.687374 | 1.671357 | 1.650515 | 1.668712 |
| 6 | 2.661374 | 2.613395 | 2.688907 | 1.416786 | 1.397752 | 1.411372 |
| 7 | 2.637658 | 2.600486 | 2.673771 | 0.803147 | 0.782514 | 0.797477 |
| 7 | 2.650468 | 2.608143 | 2.682578 | 0.857336 | 0.837601 | 0.847664 |

Feed forward backloop algorithm is used and the trained model is as shown in fig. 2

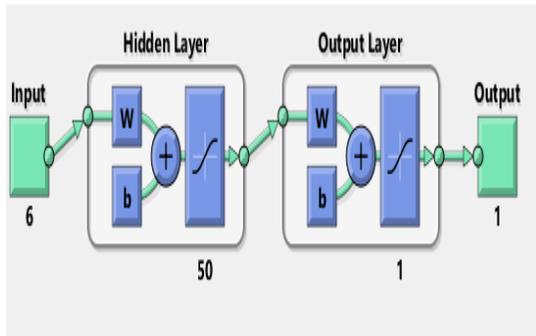

Fig 2: Trained ANN model

The weight (priority) assigned to each input can be seen from fig 2 and also the mathematical function $\Sigma X_i W_i$ +**Bias** operating can be visualised. After each summation, a sigmoid function is generated which transmits the current value to the next layer. And the regression plot showing the accuracy of training is as shown in fig 3.

From fig 3, it can be observed that the trained model fits accurately with the ideal. Hence, the training of the network is satisfactory. The trained neural network was then tested with test data. The ANN model classified each of the test data into their respective classes.

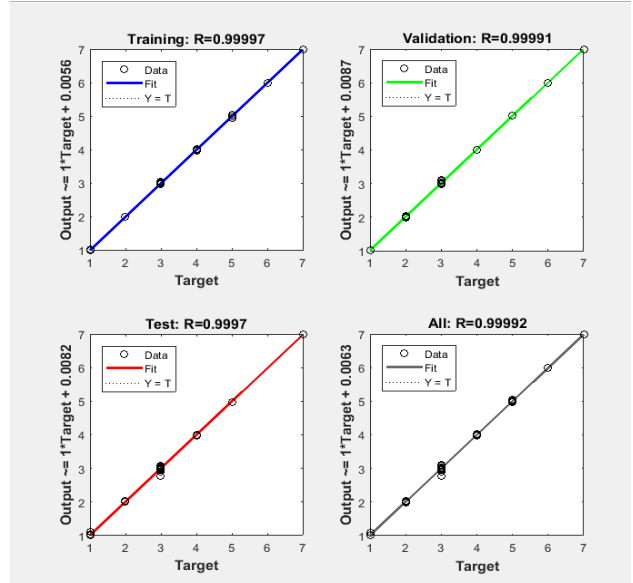

Fig 3: Regression plot of trained model

Table 2, gives the frequency of classification of each fault. The proposed ANN model classified all the test data accurately.

Table 2: Test results

| Fault | Frequency of classification |
|---|---|
| No fault (1) | 11 |
| Overload (2) | 12 |
| Ground fault (3) | 15 |
| Locked rotor (4) | 17 |
| Unbalanced voltage (5) | 3 |
| Single phasing, under voltage (6) | 3 |
| Overvoltage (7) | 5 |
| Total | 67 |

## IV. INTERFACING WITH REAL MOTOR SYSTEM

To interface the developed ANN model with the motor system, the data need to be collected continuously in real time from current and voltage sensors. The obtained values are transmitted to Wi-Fi router by using a reliable Wi-Fi connection. This data is then transferred and collected on a cloud platform (any of the many available) by using an

Ethernet cable between the router and storage system. From the cloud, the data is continuously transmitted onto the MATLAB workspace. When a fault occurs, the trained ANN model detects the fault immediately and notifies the control system to take action.

## V. SIMILAR MACHINE LEARNING BASED TECHNIQUES.

Of late, many other methodologies like Support vector machines (SVM), fuzzy logic, MCSA are being considered for applications involving predictive maintenance of electrical equipment. However, they involve lot of mathematics, are parameter based and hence, are error prone when used for large and complex systems. High mathematical dependency of these techniques also makes them not so efficient for electromechanical systems due to their inability to handle non- linearity [13]. ANN on the other hand is non-parametric, does not require complex mathematics and can be realised using simple tools and hence, is a very potent methodology for fault detection and maintenance [9].

## VI. FUTURE SCOPE

Predictive maintenance of electrical equipment, including induction motors using MACHINE LEARNING approaches is still in proposal stage in many developing countries including India. ANN due to its ability to learn from previous experiences can effectively address catastrophic failures and is perfect to handle non- linearity in data. The proposed ANN model is satisfactorily interfaced with real systems and is proven to be reliable. Hence it can serve as the most efficient and economical predictive maintenance tool in future. The proposed model can also be extended for D.C and synchronous motors.

## VII. CONCLUSION

This work depicts the potential of machine learning tool (ANN) in detecting the electrical faults of a three phase induction motor. The data collected in real time from a 1/3 HP, 208 V three phase induction motor was used to train and test the neural network. The developed neural network classifies all the test data into their respective classes with almost 100% accuracy.
The accuracy of the trained model is reflected by the results obtained using the test data and also from the regression plot shown in figure 3. In case there are errors in classification, the accuracy can be improved by increasing the number of hidden layers and by choosing optimum number of neurons. The proposed model gives accurate results and hence, prevents event progression during the occurrence of faults and protects vital electrical equipment.